\documentclass{article}

\usepackage{arxiv}

\usepackage[utf8]{inputenc} 
\usepackage[T1]{fontenc}    
\usepackage{url}            
\usepackage{booktabs}       
\usepackage{amsfonts}       
\usepackage{amsmath}
\usepackage{amssymb}
\usepackage{amsbsy}
\usepackage{newtxtext,newtxmath}
\usepackage{bm}
\usepackage{nicefrac}       
\usepackage{microtype}      
\usepackage{lipsum}		
\usepackage{graphicx}
\usepackage[round]{natbib}
\usepackage{doi}
\usepackage[printonlyused, nolist]{acronym}
\usepackage{lineno}
\nolinenumbers

\title{Dynamic Time Warping Clustering to Discover Socio-Economic Characteristics in Smart Water Meter Data}

\date{November 7, 2020}	

\author{ \href{https://orcid.org/0000-0003-2137-985X}{\includegraphics[scale=0.06]{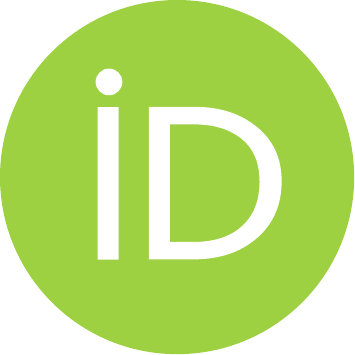}\hspace{1mm}D.B.~Steffelbauer}\thanks{Corresponding author: \texttt{david.steffelbauer@ntnu.no}}\\
	Department of Civil and Environmental Engineering \\ 
	Norwegian University of Science and Technology \\
	S.P. Andersens veg 5, 7031 Trondheim, Norway 
	\And
    {E.J.M. Blokker} \\
    Drinking Water Infrastructure Team \\
	KWR Water Cycle Research Institute \\ Groningenhaven 7, 3433 PE Nieuwegein, Netherlands\\ 
	\AND
	{\href{https://orcid.org/0000-0002-8795-1583}{\includegraphics[scale=0.06]{orcid.pdf}}\hspace{1mm}
    S.G. Buchberger} \\
    College of Engineering and Applied Science \\
    University of Cincinnati \\ 
    Cincinnati, OH 45221, United States\\ 
    \AND
	{\href{https://orcid.org/0000-0002-0335-5099}{\includegraphics[scale=0.06]{orcid.pdf}}\hspace{1mm}
    A. Knobbe} \\
    Leiden Institute of Advanced Computer Science \\ 
    Leiden University \\ 
    Niels Bohrweg 1, 2333 CA Leiden, Netherlands \\
    \AND
	{\href{https://orcid.org/0000-0003-0989-5456}{\includegraphics[scale=0.06]{orcid.pdf}}\hspace{1mm}
    E. Abraham} \\
    Department of Water Management \\
    Delft University of Technology \\
    Stevinweg 1, 2628 CN Delft, Netherlands
}

\renewcommand{\headeright}{ }
\renewcommand{\undertitle}{Preprint of Article Published in Journal of Water Resources Planning and Management,\\ Vol. 147 Issue 6 (June 2021), \protect{\url{https://doi.org/10.1061/(ASCE)WR.1943-5452.0001360}}.}
\renewcommand{\shorttitle}{\textit{Steffelbauer et al.} (2021): DTW Clustering to Discover Socio-Economic Characteristics in SWM Data}

\hypersetup{
pdftitle={Dynamic Time Warping Clustering to Discover Socio-Economic Characteristics in Smart Water Meter Data},
pdfsubject={cs.LG},
pdfauthor={D.B.~Steffelbauer, E.J.M.~Blokker, S.G.~Buchberger, A.~Knobbe, E.~Abraham},
pdfkeywords={Water demand, Water meters, Hydrologic data, Water supply systems, Hydraulic models, Social factors, Economic factors},
}

\begin{document}
\maketitle
\begin{abstract}
Socio-economic characteristics are influencing the temporal and spatial variability of water demand – the biggest source of uncertainties within water distribution system modeling.
Improving our knowledge on these influences can be utilized to decrease demand uncertainties.
This paper aims to link smart water meter data to socio-economic user characteristics by applying a novel clustering algorithm that uses dynamic time warping on daily demand patterns.
The approach is tested on simulated and measured single family home datasets.
We show that the novel algorithm performs better compared to commonly used clustering methods, both,  in finding the right number of clusters as well as assigning patterns correctly.
Additionally, the methodology can be used to identify outliers within clusters of demand patterns.
Furthermore, this study investigates which socio-economic characteristics (e.g. employment status, number of residents) are prevalent within single clusters and, consequently, can be linked to the shape of the cluster’s barycenters. 
In future, the proposed methods in combination with stochastic demand models can be used to fill data-gaps in hydraulic models.
\end{abstract}

\keywords{Water demand \and Water meters \and Hydrologic data  \and Water supply systems  \and Hydraulic models  \and Social factors  \and Economic factors}

\section{Introduction}
Water utilities make use of hydraulic simulation software to design and operate their systems in a more effective way \citep{Haestad2003}.
However, models of \acp{wds} consist of thousands or tens of thousands of parameters (the length, diameter and roughness of every pipe or the water demand at every node). 
These parameters are mostly unknown, have to be estimated through model calibration \citep{Zhou2018} and are fraught with uncertainties. 
Especially, water demand plays a crucial role in the dynamics of \acp{wds}, since it fluctuates over a variety of temporal and spatial scales depending on the type of consumers \citep{Hutton2012,Diaz2020}. 
Additionally, due to the low density of metered consumers and the difficulty to obtain large amounts of demand data in real-time, the  variability of water demand is the biggest source of uncertainty in \ac{wds} modeling.

Over the last decade, \acp{swm}  that measure and transmit water consumption data at single household level are available in high temporal resolution, from sub-second up to one hour \citep{Boyle2013}, potentially overcoming limitations of current  metering practices. 
These devices have the potential to revolutionize \ac{wds} modeling \citep{Gurung2014,Nguyen2018,Stewart2018}. 
However, the large-scale roll-out of \acp{swm} globally is yet to happen since technology adoption barriers ---caused by financial, cyber security, privacy issues--- hinder the wide-spread deployment of this new technology \citep{Cominola2015}. 
Furthermore,  for water utilities adopting this new technology, the  cost-benefit trade off has not yet been quantitatively justified~\citep{Cominola2018,Monks2019}. 
Besides data management challenges associated with big data streams~\citep{Shafiee2020}, water companies are further challenged to generate relevant knowledge from the raw consumption data that has to be useful for their hydraulic computer models.
However, a combination of system wide consumer information and data from a few \acp{swm} represents a promising approach to reduce modeling uncertainties, by filling in data gaps at the unmeasured locations according to their consumers' characteristics, without extensively measuring real-time water demand at every node in a \ac{wds}.
The question is: Is it possible to link raw \ac{swm} data to rather general consumer information?

Deriving valuable information from \acp{swm} is by far not trivial, since water use is stochastic by nature \citep{Blokker2010}; no one operates water end-use devices (shower, water tap, toilet, \ldots) exactly at the same time each day and extracts precisely the same amount of water during each usage.
Nevertheless, by building periodic means over a certain number of days, patterns in water usage behavior emerge. 
These patterns are called \textit{daily demand patterns}. 
The patterns contain information about consumer's daily routines, reveal irregular consumption behaviors  and are shaped by their socio-economical characteristics, e.g., age, gender, economic situation, employment status or family composition. Hereinafter, we will refer to this information as the underlying \textit{high-level information}.
This paper will show that data mining techniques can be used to automatically distinguish daily demand patterns into groups according to their underlying high-level consumer information.

We aim to answer following two questions by applying a novel clustering algorithm on daily demand patterns generated from \ac{swm} data:
\begin{enumerate}
	\item[Q.1.] How many distinct daily demand patterns are  in a specific \ac{swm} dataset?
	\item[Q.2.] Can we draw conclusions from these pattern shapes on the  consumers' underlying high-level information?
\end{enumerate}
The  proposed methods are tested on two artificial   \ac{swm} datasets generated with the stochastic demand modeling software SIMDEUM \citep{Blokker2010} and a real-world \ac{swm} dataset from Milford, Ohio \citep{Buchberger2003}.

\subsection{Related work}
Cluster analysis belongs to the family of  unsupervised learning algorithms and is a technique to find groups in datasets \citep{Lin2012}.  
For \ac{swm} data analysis, clustering can be used to segment users into groups with similar water-use behaviour~\citep{Cominola2019}, e.g., commercial vs. residential or single households vs. multi-family homes. 
While customer segmentation was mostly focusing  in the past on smart meters that measure energy consumption~\citep{Espinoza2005,Nizar2009,Nambi2016,Kwac2014}, only few studies applied customer segmenation on \ac{swm} data. 
Most research uses $k$-Means clustering \citep{Lloyd1982} in combination with different distance measures. 
\citet{McKenna2014} clustered \ac{swm} data  into commercial and residential patterns by applying  $k$-Means on features extracted through fitting Gaussian mixture models. 
\citet{Mounce2016} used $k$-Means++ \citep{Arthur2007} with correlation distance to cluster data in residential and commercial groups. 
\citet{Garcia2015} classified demand patterns using $k$-Means clustering.
\citet{Cheifetz2017} made use of Fourier-based time series models for clustering demand patterns. 
The patterns were qualitatively interpreted as residential, commercial, office, industrial or noise. 
\citet{Cardell-Oliver2016} identified groups of similar households by features of their high-magnitude water use behaviours based on previous work~\citep{Cardell-Oliver2013a,Cardell-Oliver2013,Wang2015}.
\citet{Cominola2018} applied customer segmentation analysis simultaneously on water and electricity data by clustering extracted eigen-behaviors and linked the clusters to a list of user psycho-graphic features.
Recently, \citet{Cominola2019} coupled non-intrusive end-use disaggregation with customer segmentation to identify and cluster primary water use behaviors.

Clustering techniques are  also used as a prior step to demand forecasting. 
For example, \citet{Aksela2011} constructed clusters with $k$-Means according to their average weekly consumption before forecasting future  demand. 
\citet{Candelieri2017} clustered \ac{swm} data using $k$-Means with cosine distance, first to split data into weekdays and weekends, then to split the data into residential, non-residential and mixed type clusters. 

In contrast to the studies mentioned above, this paper employs \ac{sdtw} as time series clustering distance measure.
This distance measure is capable of optimally aligning two sequences in time by non-linearly warping the time-axes of the sequences until their dissimilarity is minimized \citep{Durrenmatt2013}. 
The time when people use water is highly variable among different users with otherwise similar socio-economic characteristics~\citep{Blokker2008}. 
The \ac{sdtw} distance measure can expose similarities in daily schedules of inhabitant's water use that are shifted in time, which are not detectable by using linear time distance measure (Euclidean, correlation).
Originally developed for speech recognition \citep{Sakoe1978}, \ac{dtw} has been applied in the field of water management in the past, but never in the context of time series clustering.
Past applications of \ac{dtw} in water management included burst detection \citep{Huang2018}, analysing residence times in waste water treatment plant reactors \citep{Durrenmatt2011}, sewer flow monitoring \citep{Durrenmatt2013} or identifying water demand end-uses \citep{Yang2018,Nguyen2014}. 
\subsection{Contributions}
Although clustering of \ac{swm} data has been done before, this paper's approach is innovative in many aspects. 
First, a novel method is proposed to cluster \ac{swm} data that is capable of finding similarities in daily demand patterns even if they are shifted in the time domain. 
Hence, it should outperform clustering methods with fixed time distance measures (e.g. Euclidean~\citep{Mounce2016,Garcia2015}).
Second, the proposed methods are tested on \ac{swm} datasets that were simulated with the stochastic demand modeling software SIMDEUM. 
Since the ground truth of these datasets are known, it enables measuring and comparing the performance of different clustering approaches. 
Third, while former work focused on identifying different types of consumer classes by mainly distinguishing between residential and commercial use, this work goes one step further by investigating which underlying high-level information of residential customers is prevalent in different clusters, e.g. by looking at work schedules or the number of household residents. 

Furthermore, we want to highlight the simplicity of the proposed approach compared to other methods. 
While most of the discussed studies used clustering on burdensome obtained surrogate parameters (e.g., eigenbehaviours~\citep{Cominola2019}, high-magnitude water use behaviors~\citep{Cardell-Oliver2016}, parameters from fits from Gaussian distributions~\citep{McKenna2014} or Fourier regression mixture models~\citep{Cheifetz2017}), the \ac{sdtw} clustering method is applied directly on the demand patterns and, hence, does not risk losing valuable information contained in the raw data. Additionally, \ac{sdtw} enables user segmentation using water consumption data  without the need for additional information as, for example, electricity~\citep{Cominola2018} or end-use disaggregated water consumption data~\citep{Cominola2019}. Moreover, the \ac{sdtw} clustering approach is an unsupervised algorithm with no need for prior information nor previous calibration of, for example, consumption threshold parameters.
\section{Materials and Methods}
\subsection{Water demand pattern generation}\label{sec:met_data_preparation}
\ac{swm} data (and demand patterns) are time series \cite{Shumway2010}. 
A time series $\mathbf{x}$ of length $M$ is a sequence of data points in strict chronological order
\begin{linenomath*}
	\begin{equation}
	\mathbf{x} \ = \ \left\{x_t\right\} \ = \ (x_1, x_2, \ldots, x_M) \ \in \mathbb{R}^M .
	\end{equation}
\end{linenomath*}
A water demand pattern $\overline{\mathbf{x}}_P$ is generated by building periodic means from a \ac{swm} time series $\mathbf{x}$
\begin{linenomath*}
	\begin{equation}
	\overline{\mathbf{x}}_{j}^{P} \ = \ \frac{1}{N_P} \sum_{i=1}^{N_P} x_{P(i-1) + j} \quad \forall j = 1, 2, \ldots ,P , \label{eq:periodic_mean}
	\end{equation}
\end{linenomath*}
where $P$ is the period length and $N_P = \left\lfloor \frac{M}{P} \right\rfloor $ is the number of full periods in $\mathbf{x}$. 
More specifically, we will deal throughout this paper with daily demand patterns($P=24$ hours).
\subsection{\acf*{sdtw}}
Unlike Euclidean distance, \ac{sdtw} is able to compare time series of variable size and is robust to shifts or dilations across the time dimension \cite{Cuturi2017}. 
The \ac{sdtw} method computes the best possible alignment between time series. 
This is relevant for \ac{swm} data since the daily water use behaviours of different households might be similar, but shifted in the time domain due to different daily schedules, e.g. caused by different wake-up, working  or commuting times.
\subsubsection{Two time series}
Let $\mathbf{x}  \in \mathbb{R}^M$ and $\mathbf{y} \in \mathbb{R}^N$ be two time series. 
Note that $\mathbf{x}$ and $\mathbf{y}$ do not have to be equally long, nor do they have to have the same sampling rate. 
Since this publication focuses on clustering daily demand patterns, the time series are always of the same length and sampling rate introduced by the periodic mean in  Eq.\eqref{eq:periodic_mean}. 
First, the elements of a pairwise distance matrix $D  \in \mathbb{R}^{M \times N}$ are computed between the points of two time series $\mathbf{x}$ and $\mathbf{y}$ 
\begin{linenomath*}
	\begin{equation}
	D_{mn} \ = \ \delta \left( x_m, y_n \right) , \label{eq:distance_matrix}
	\end{equation}
\end{linenomath*}
where $\delta$ is an arbitrary distance measure. 
A path connecting the upper left corner and the right bottom corner of $D$ that only allows moves to the right ($\rightarrow$), diagonal ($\searrow$) or down ($\downarrow$)  is called a warping path $\mathbf{p}$. 
This path is used  to align the two time series $\mathbf{x}$ and $\mathbf{y}$. 
The warping path $\mathbf{p}$ is linked to the binary alignment matrix $A \in \mathbb{R}^{M \times N}$: 
\begin{linenomath*}
	\begin{equation}
	\mathbf{p} \ = \ \left( \rightarrow, \searrow, \downarrow, \rightarrow, \dots  \right) 
	\quad \triangleq \quad 
	A \ = \ \begin{bmatrix}
	1 & 1 & 0 & 0 & \cdots & \\
	0 & 0 & 1 & 0 & 		  & \\
	0 & 0 & 1 & 1 & 		  & \\
	\vdots &  &  &  & \ddots & \\
	&  &  &  &  & 1\\
	\end{bmatrix} . \label{eq:warping_path}
	\end{equation}
\end{linenomath*}
In the following, we will write $\aleph \in \{ 0, 1 \}^{M \times N}$ as the set of all possible (binary) alignment matrices $A$.
The warping distance $d$ along a warping path $\mathbf{p}$  is defined through
\begin{linenomath*}
	\begin{equation}
	d(A, D) \ = \  \sum_{ij} A_{ij} D_{ij} \ = \ \text{Tr} \left(A^T D\right) ,
	\end{equation}	
\end{linenomath*}
where $\text{Tr}(\cdot)$ is the trace of a matrix. 
The optimal warping path $\mathbf{p}^\ast$ with minimal distance $d^\ast$ is computed with  \ac{sdtw} in the following way \cite{Cuturi2017}
\begin{linenomath*}
	\begin{equation}
	d^\ast(\mathbf{x}, \mathbf{y}) \ = \min_{ A \in \aleph } \left( d(A, D) \right)
	\,  = \, - \log \left(\sum_{A \in \aleph } e^{  - {d(A, D)} }\right). \label{eq:dtw}
	\end{equation}
\end{linenomath*}
The \ac{sdtw} distance measure integrates over all possible alignments, is differentiable and leads to a robust smooth solution in an optimization framework \cite{Cuturi2017}. 
Although the set of all possible alignment matrices $\aleph$ grows exponentially with $M$ and $N$, Eq.\eqref{eq:dtw} can be recursively solved with computational complexity of order $\mathcal{O} \left(M  N\right)$ starting from $r_{0,0}=0$:
\begin{linenomath*}
	\begin{equation}
	r_{i,j} \ = \ \delta\left(x_i, y_j\right) + \min{} \left( r_{i-1, j-1}, r_{i-1, j}, r_{i, j-1}  \right) . \label{eq:dtw_recursion}
	\end{equation}
\end{linenomath*}
\subsubsection{Multiple time series}
The optimal distance $d^\ast$ can be used  to  average over multiple time series. 
The ability to build such averages is a necessary condition for time series clustering.
Let $\{\mathbf{y}_l\} = (\mathbf{y}_1, \ldots, \mathbf{y}_L)$ be a family of $L$ time series. To average $\{\mathbf{y}_l\}$ with \ac{sdtw} , the following minimization problem has to be solved
\begin{linenomath*}
	\begin{equation}
	\min_{\mathbf{x^\ast}  \in \mathbb{R}^M} \sum_{i=1}^{L} \  d^\ast(\mathbf{x^\ast}, \mathbf{y}_i) . \label{eq:dtw_averaging}
	\end{equation}
\end{linenomath*}
This problem is solved using a quasi-Newton method, the Broyden-Fletcher-Goldfarb-Shanno (BFGS) algorithm \cite{Nocedal2006}.
The solution  $\mathbf{x^\ast}$ is called the barycenter of $\{\mathbf{y}_l\}$ .
\subsection{Clustering}
The $k$-Means clustering method  \cite{Lloyd1982} is used to identify different demand patterns in the  \ac{swm} data. 
The principle behind $k$-Means algorithm is to separate the data into a preset number of $k$ clusters ${\bm{\mathcal{C}} = \left\{C_1, \ldots, C_k \right\}}$ that minimize intra-cluster variability and maximize inter-cluster differences \cite{McKenna2014}. 
Let $\{\mathbf{y}_l\}$ be again a family of $L$ time series.
Then $k$-Means clustering in the Euclidean metric equals minimizing the following nested sums
\begin{linenomath*}
	\begin{equation}
	\min_{\bm{\mathcal{C}}} \sum_{j=1}^{k}  \sum_{\mathbf{y} \in C_j}  ||\mathbf{y} - \bm{\mu}_j||_2 ,\label{eq:kmeans_euclidean}
	\end{equation}
\end{linenomath*}
where $\bm{\mu_j}$ is the within-cluster mean of $C_j$. 
Note that the Euclidean norm $|| \, \cdot \,||_2$ is only valid when the time series are of equal length ($M$=$N$).  
Analogously, clustering using the \ac{sdtw} distance measure  is defined as
\begin{linenomath*}
	\begin{equation}
	\min_{\bm{\mathcal{C}}} \sum_{j=1}^{k}  \sum_{\mathbf{y} \in C_j}   d^\ast (\mathbf{x}_{j}^\ast, \mathbf{y}) , \label{eq:kmeans_dtw}
	\end{equation}
\end{linenomath*}
where $\mathbf{x}_{j}^\ast$ is the barycenter of the $j$-th cluster. 

Furthermore, we introduce a simplified version of the $k$-Means clustering algorithm. 
Instead of investigating the whole daily demand time series $\mathbf{y}$, this algorithm uses only the mean ($E\left[\mathbf{y}\right]$) and the standard deviation ($\sqrt{E[\mathbf{y}^2] - \left( E\left[\mathbf{y}\right] \right)^2}$) of the daily pattern during work hours (10:00-16:00). 
Hence, the feature space is reduced to a two-dimensional space through following transformation
\begin{linenomath*}
	\begin{equation}
		\mathbf{y} \ \Rightarrow \ \begin{pmatrix}
			E\left[\mathbf{y}\right] \\  
			\sqrt{E[\mathbf{y}^2] - \left( E\left[\mathbf{y}\right] \right)^2}
		\end{pmatrix}\label{eq:simple_clustering} .
	\end{equation}
\end{linenomath*}
Within this work the problem defined in Eq.$\eqref{eq:kmeans_euclidean}$ will be called Euclidean clustering, Eq.$\eqref{eq:simple_clustering}$ Simple clustering, and Eq.$\eqref{eq:kmeans_dtw}$ \ac{sdtw} clustering. 
The latter is capable of finding more general similarities in patterns by allowing more freedom in the time domain and will be benchmarked against  Simple and Euclidean clustering.
The initial cluster centers are seeded according to the $k$-Means++ algorithm \cite{Arthur2007} to increase the method's robustness. 
Prior to clustering, the time series $\mathbf{y}_i$ can be normalized:
\begin{linenomath*}
	\begin{equation}
	\mathbf{y}_{i}^{\prime} \ = \ \frac{\mathbf{y}_i - \min(\mathbf{y}_i) }{\max(\mathbf{y}_i) - \min(\mathbf{y}_i) } . \label{eq:normalization}
	\end{equation}
\end{linenomath*}
\subsection{Performance Measures}
Success and error rates are used to validate the clustering results based on the ground truth~\cite{Witten2011}, whereas Silhouette coefficients are used to validate the clusters based on the (dis)similarities of their members~\cite{Rousseeuw1987}.
Two cases have to be distinguished for validating clustering results: (i) if the ground truth is known; (ii) if there exist no information about the true nature of the outcomes. 
For the first case, the correct allocation of distinct patterns is known and  one can compute a success respectively an error rate   \cite{Witten2011}.
In the second case, the allocation and the number of distinct patterns is unknown. 
For  datasets with unknown ground truth, the clustering results can still be validated based on Silhouette coefficients.
\subsubsection{Success and error rate}
\ac{tp} is the case if a pattern is assigned to the correct cluster. 
\ac{fp} means that a  pattern belonging to another cluster is wrongly assigned to the current cluster. 
A \ac{tn} is the case when a pattern from another cluster is correctly assigned to the other cluster. 
Finally, a \ac{fn} is the case when a pattern belonging to the cluster is wrongly assigned to another cluster. 
One can define an overall \ac{sr} with all above mentioned cases through following equation \cite{Witten2011}
\begin{linenomath*}
	\begin{equation}
	\text{SR} \ = \ \frac{\text{TP} + \text{TN}}{\text{TP} + \text{TN} + \text{FP} + \text{FN}} . \label{eq:success_rate}
	\end{equation}
\end{linenomath*}
The \ac{er} is the complement of \ac{sr}:
\begin{linenomath*}
	\begin{equation}
	\text{ER} \ = \ 1 - \text{SR} . \label{eq:error_rate}
	\end{equation}
\end{linenomath*}
\subsubsection{Silhouette Coefficients}
Silhouette coefficients $S_i$ are properties of a single time series $\mathbf{y}_l$ \cite{Rousseeuw1987}.  
They can be used to determine the quality of clusters when the ground truth is unknown. 
The $S_i$ values are computed as a combination of two scores: (i) the mean intra-cluster distance and (ii) the distance between a sample and the nearest cluster that $\mathbf{y}_l$ is not part of. 
The mean intra-cluster distance is defined as the average distance of time series $\mathbf{y}_l$ to all other time series $\mathbf{y}_j$ that are members of the same cluster $C_i$. 
Let $\mathbf{y}_l$ be the $l^{\text{th}}$ member of the time series belonging to cluster $C_i$, then its intra-cluster distance (the mean distance between all members $\mathbf{y}_j$ of cluster  $C_i$) $a(\mathbf{y}_l)$ is defined as
\begin{linenomath*}
	\begin{equation}
	\ a(\mathbf{y}_l) \ = \ \frac{1}{| C_i| - 1} \sum_{\substack{\mathbf{y}_j \in C_i \\ j \neq l}} d(\mathbf{y}_j, \mathbf{y}_l) .
	\end{equation}
\end{linenomath*}
$|C_i|$ is the number of samples in cluster $C_i$ and $d$ is an arbitrary distance measure. 
The second score $b(\mathbf{y}_l)$ is the distance $d$ between the time series $\mathbf{y}_l$ of $C_i$ and its nearest cluster $C_{j}^\ast$: 
\begin{linenomath*}
	\begin{equation}
	b(\mathbf{y}_l) \ := \ \min_{i \neq j} \frac{1}{|C_j|} \sum_{\mathbf{y}_k \in C_j} d(\mathbf{y}_k, \mathbf{y}_l) \ = \  
	\frac{1}{|C_{j}^\ast|} \sum_{\mathbf{y}_k \in C_{j}^\ast} d(\mathbf{y}_k, \mathbf{y}_l) ,
	\end{equation}
\end{linenomath*}
where $\mathbf{y}_k \in C_j$ represents the members of the cluster $C_j$.
The two scores are combined in following way to obtain the Silhouette coefficient of a time series $\mathbf{y}_l$ \cite{Rousseeuw1987}:
\begin{linenomath*}
	\begin{equation}
	S(\mathbf{y}_l) \ = \frac{b(\mathbf{y}_l) - a(\mathbf{y}_l)}{\max \left(a(\mathbf{y}_l), b(\mathbf{y}_l)\right)} . \label{eq:silhouette}
	\end{equation}
\end{linenomath*}
By definition, the Silhouette coefficient is between $-1 \leq S_i \leq 1$. 
Higher $S_i$ values relate to a model with better defined clusters, i.e., each time series is closer to its own cluster members than to the nearest cluster.

The specific number of clusters $k$ is required as an input parameter for the $k$-Means algorithm.
Average Silhouette coefficient for all $L$ time series can be used to compare clustering results for different $k$ to decide on the number of clusters that are in the data
\begin{linenomath*}
	\begin{equation}
	\overline{S} \ = \ \frac{1}{L} \sum_{l=1}^{L} S(\mathbf{y}_l) . \label{eq:average_silhouette}
	\end{equation} 
\end{linenomath*}
Investigating the behavior of $\overline{S}$ as a function of $k$ is subsequently called \textit{cluster analysis}.
\subsection{Datasets}
The \ac{sdtw} methodology is tested on three datasets of water use at single family homes: (i) an artificial \ac{swm} dataset generated with SIMDEUM \cite{Blokker2010} consisting of single-person households, (ii) another SIMDEUM dataset with multiple-person homes, and (iii) a measured  \ac{swm} dataset from Milford, Ohio \cite{Buchberger2003}. 
Daily demand patterns with a  time resolution of 30 minutes are generated and smoothed with a two-hour moving average. 
A more detailed description of the datasets can be found in the supplemental materials.

\subsubsection{SIMDEUM single-person households}
SIMDEUM is a water demand end-use model that is capable of simulating water usage at household level with a time resolution of down to one second \cite{Blokker2010}. 
The model generates randomly water end-use events based on statistical information of users and end-use devices. 
The information includes census data for the number of residents in a household, their age distribution, the average number of appliances, their daily routines, as well as frequency, duration and intensity for different end-uses such as kitchen tap uses, toilet flushes, number of showers taken per day, washing machine or dishwasher uses, etc.

To generate the first simulated dataset, we  simulated 100 single-person households with different daily routines. 
For each household, the water consumption of its residents for a period of 100 days was simulated.
The dataset consists of adult inhabitants \textit{with} (50) and \textit{without} (50) jobs away from home. 
Consequently, the employment status of the occupants is the underlying high-level information responsible for the different pattern shapes.
Throughout this paper, we will refer to the first half of the household demand patterns as \textit{work} patterns and to the second half as \textit{home} patterns.

\subsubsection{SIMDEUM multiple-person households}
The second simulated dataset was produced by generating SIMDEUM simulations for 200 multi-person homes with one to five residents according to the household statistics in The Netherlands (see Table 2 in \cite{Blokker2010}); again simulated for 100 days for each household.
The high-level information consists of the type of the household (one-person, two-person, family), the number of residents, their age distribution, their daily schedules, their profession (employed, unemployed, retired, kid, teenager). 

\subsubsection{Measured SWM data from Milford (Ohio)}
The measured \ac{swm} dataset contains data from \ac{swm} installed at 21 single-family houses in Milford (Ohio, USA) recorded between the $1^{st}$ of April and $31^{st}$ of October 1997 \cite{Buchberger2003}.
Since user behaviour and, hence, daily demand patterns can differ significantly between weekends and weekdays~\cite{Alvisi2007}, patterns were generated separately for  weekdays and  weekends. The underlying high-level information contains the number of residents and the pattern type (home, work).
\section{Results}

\subsection{Simulated dataset with single-person households}
\begin{figure}[htbp]
	\centering
	\includegraphics[width=\textwidth]{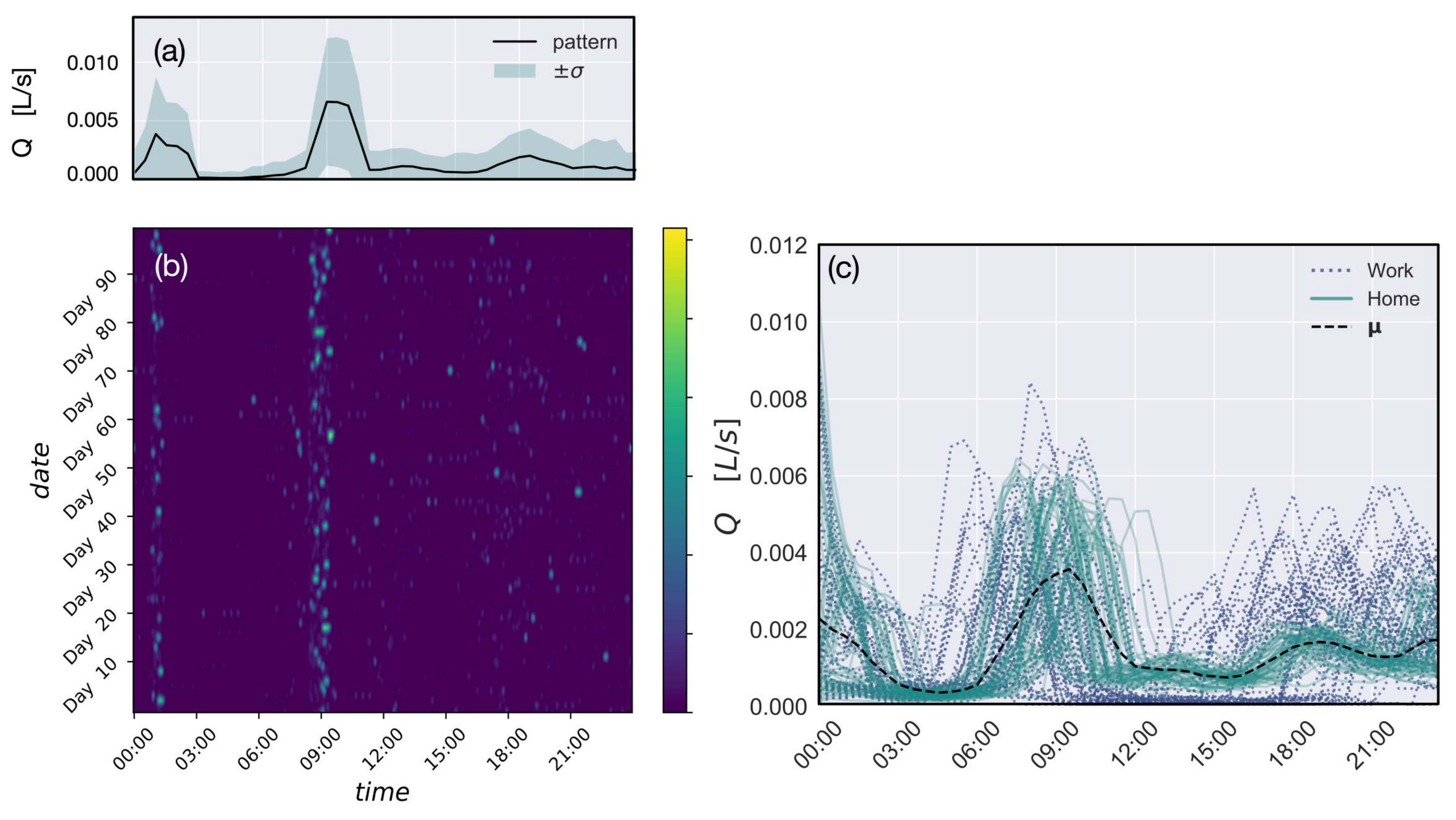}
	\caption{Representation of the SIMDEUM single-person household dataset. (a): Daily demand pattern constructed from 100 days of \ac{swm} data and its standard deviation  $\sigma$ of an example household. (b): The example household's water consumption over 100 days. 
		(c): Demand patterns (mean over 100 days) of the whole dataset showing 50 work patterns, 50 home patterns and the mean over all demand patterns $\bm{\mu}$.}\label{fig:simdeum_pattern}
\end{figure}

\begin{figure}[htb]
	\centering
	\includegraphics[width=\textwidth]{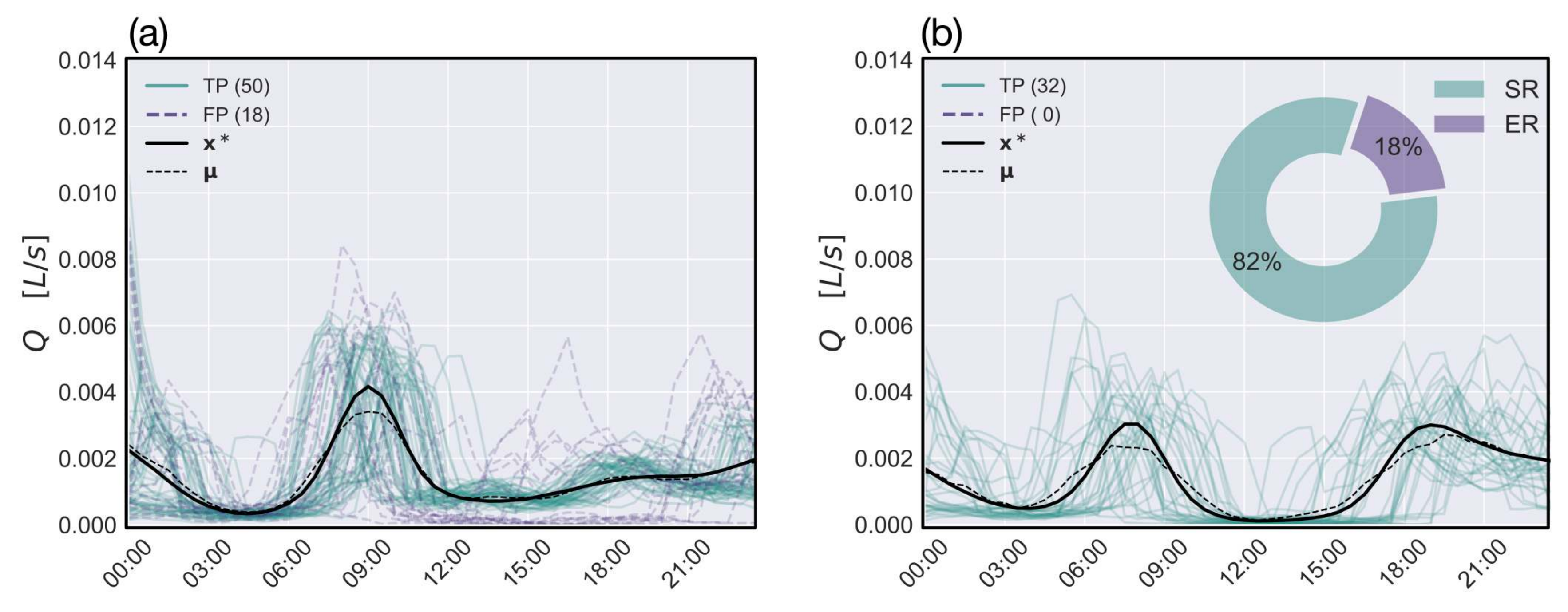}
	\caption{Clustering results for SIMDEUM single-person household dataset showing the home (a) and work cluster (b) including a doughnut chart showing the SR and ER.}\label{fig:simdeum_clustering}
\end{figure}

\begin{figure}[htbp]
	\centering
	\includegraphics[width=1.0\linewidth]{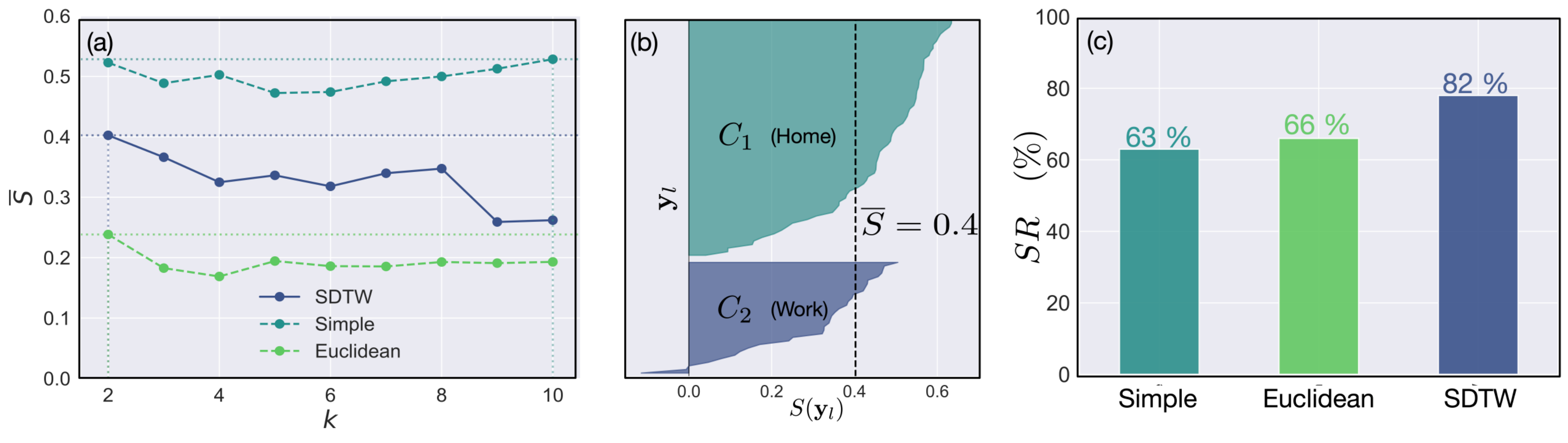}
	\caption{Cluster analysis plots for SIMDEUM single-person household dataset. (a) Average silhouette coefficient $\overline{S}$ as a function of the  number of clusters $k$ for \ac{sdtw} clustering and the benchmark algorithms. Colored dashed lines highlight the maximum $\overline{S}$ values. (b) Silhouette plot for the \ac{sdtw} measure for $k$=2 showing the silhouette coefficients of each member of the two clusters $C_1$ (home) and $C_2$ (work). (c) Comparison of the performance between \ac{sdtw} clustering and the benchmark algorithms with respect to the \acp{sr}.}\label{fig:simdeum_cluster_analysis}
\end{figure}

The dataset consists of  \textit{work} and \textit{home} patterns.
We will refer to a cluster consisting predominantly of work patterns as \textit{work cluster}, and to a cluster whose majority patterns are  home patterns as \textit{home cluster}.
The intention of this numerical experiment is to see (i) if the   \ac{sdtw} clustering approach is able to extract the employment status of the residents; (ii) if we are able to identify the correct number of distinct patterns in the dataset with cluster analysis; and (iii) how \ac{sdtw} clustering performs compared to the benchmark algorithms (Simple and Euclidean clustering).
Figure \ref{fig:simdeum_pattern} presents the SIMDEUM dataset. 
Figure~\ref{fig:simdeum_pattern}~(a) shows  the daily demand pattern  of a single-person household where  the occupant is staying at home throughout the day. 
Additionally, the standard deviation $\sigma$  is shown.  
Figure~\ref{fig:simdeum_pattern}~(b) provides the 100 days water usage data of this household.
Figure~\ref{fig:simdeum_pattern}~(c) presents the demand patterns of the whole SIMDEUM dataset (100 households). 
The work patterns are shown as dotted purple lines, the home patterns are shown as solid green lines. 
Furthermore, the mean over all patterns is shown as a dashed black line. 
Variations of the patterns from the mean are clearly visible, both, in time and  in magnitude. 
First, \ac{sdtw} clustering is applied on the dataset. 
The patterns are normalized prior to clustering to suppress the influence of different average consumption, so that the algorithm focuses only on pattern shapes and not on magnitudes.
Figure \ref{fig:simdeum_clustering}~(a) and (b) present the clustering results for $k$=2 and the performance measures; the home cluster on the left side (a) and the work cluster on the right side (b).
Demand patterns classified correctly (\ac{tp}) are shown as solid green lines, \ac{fp} are depicted as dashed purple lines.
The work cluster is a pure cluster consisting only of work patterns. 
The \ac{sr} is $82 \%$ and $\ac{er}$ is 18 \% (see the doughnut chart in Figure \ref{fig:simdeum_clustering} (b)).
Furthermore, Figure \ref{fig:simdeum_clustering} (a) and (b) show the barycenters $\mathbf{x}^\ast$.
Additionally, the within-cluster mean $\bm{\mu}$ is shown to illustrate the difference of  $\mathbf{x}^\ast$ and $\bm{\mu}$.
It can be seen that the \ac{sdtw} clustering approach is capable of segmenting the daily demand patterns according to the employment status of the inhabitants. 
Furthermore, the barycenters show the expected water use behavior of the user groups. 
The users within the home cluster use water over the whole day, while users in the work cluster have almost zero consumption while their residents are at work. 
Second, we perform a cluster analysis to see if we can identify the correct number of patterns. 
Note that the distinct number of patterns is two (work, home), since there is no other high-level information contained in the data.
The results of the cluster analysis with the \ac{sdtw} method are presented in Figure \ref{fig:simdeum_cluster_analysis} (a) and compared with the benchmark algorithms.
Additionally, the individual Silhouette coefficients $S(\mathbf{y}_l)$ are shown for the correct number of clusters $k$=2  (Figure~\ref{fig:simdeum_cluster_analysis} (b)). 
The maximum $\overline{S}$ value indicates the most probable number of clusters in the dataset. 
The \ac{sdtw} and the Euclidean clustering approach are capable of finding the correct number of clusters (maximum at $k$=2), whereas the simple algorithm overestimates the number of clusters. 
Figure \ref{fig:simdeum_cluster_analysis} (c) shows a comparison of the three clustering algorithms with respect to the \acp{sr}, where the \ac{sdtw} algorithm clearly performs best.
\subsection{Simulated dataset with multiple-person households}
\begin{figure}[htbp]
	\centering
	\includegraphics[width=1.0\linewidth]{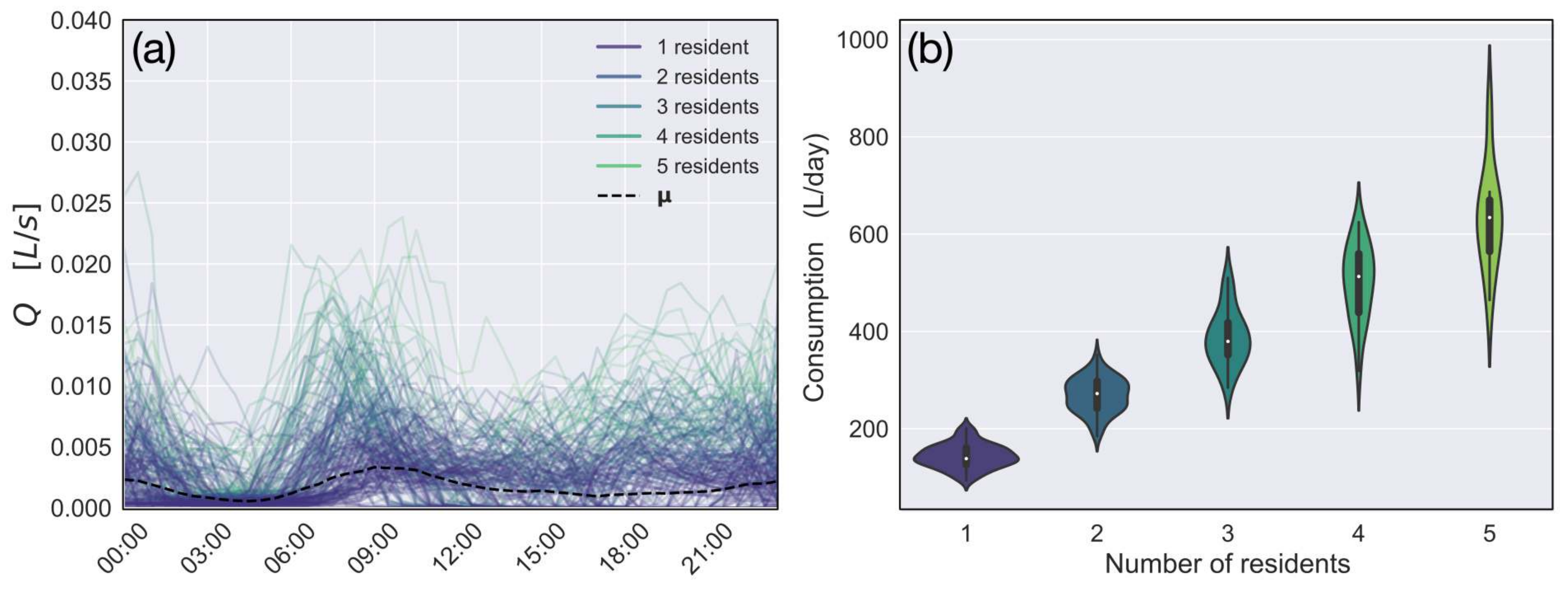}
	\caption{Representation of 200 simulated multi-person households. (a): Daily demand pattern of the households colored according to the number of residents. The mean $\bm{\mu}$ is depicted as a dashed black line. (b): Violin plots for the average daily consumption versus  the  number of residents.}
	\label{fig:simdeum2_dataset}
\end{figure}

The purpose of this experiment is to apply the \ac{sdtw} algorithm on a more complex dataset. 
Figure~\ref{fig:simdeum2_dataset} shows the dataset. 
On average, the consumption grows linearly with the household size or the type of household (one-person, two-person, family).
Hence, this high-level information (the type of households, the number of users) is supposed to be influential  on the pattern shape.
Furthermore, the growing variance in the data leads to a lot of consumption overlaps between households of different resident numbers, making it difficult to segment the data by consumption only.
First, we will perform a cluster analysis to identify the number of distinct patterns in the dataset, followed by a closer look on the cluster's barycenters. 
Second, we will try to identify the most influential high-level information.
Again, the performance of the \ac{sdtw} method is compared with Simple and Euclidean clustering.

\begin{figure}
	\centering
	\includegraphics[width=\textwidth]{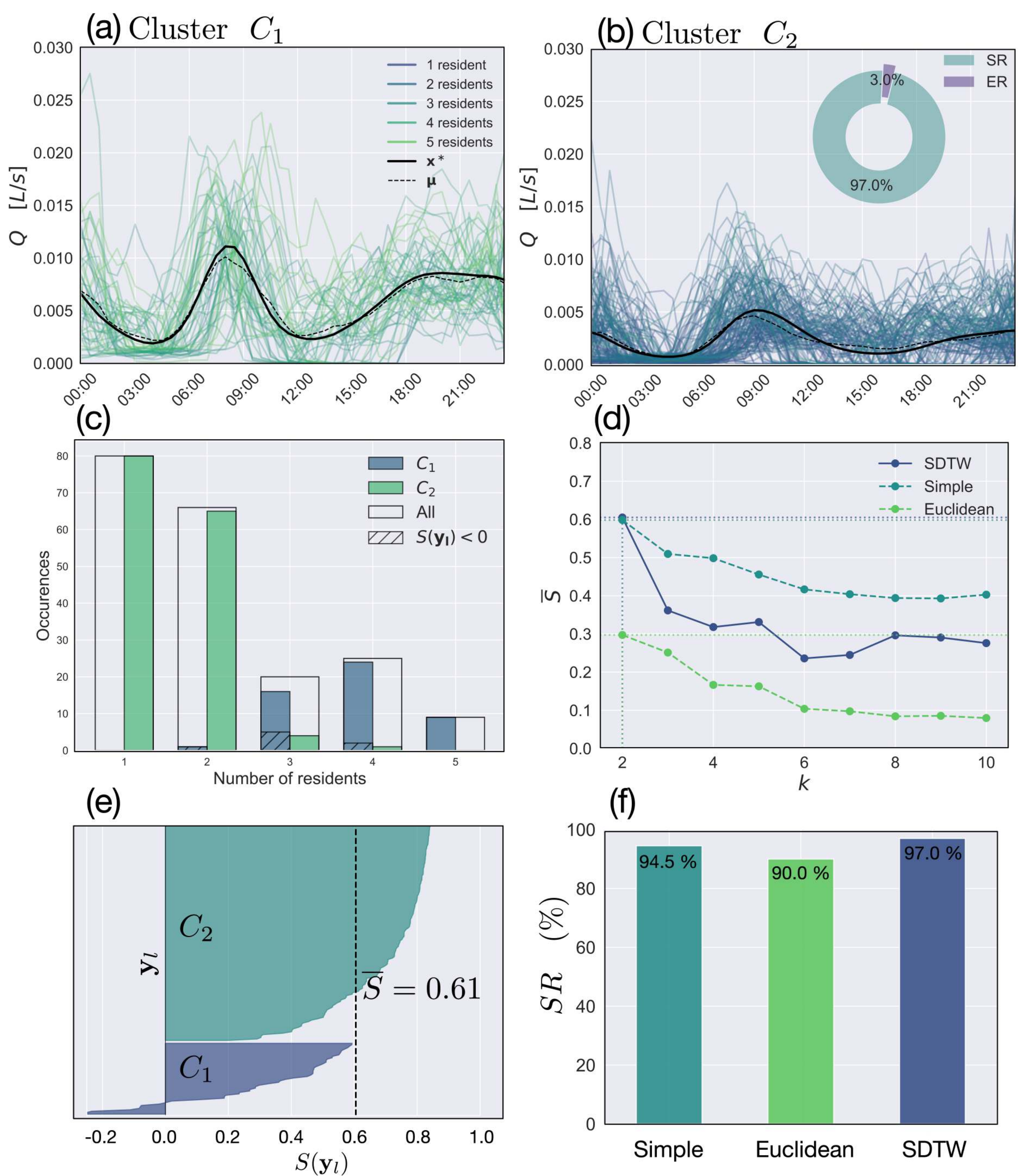}
	\caption{Clustering results for SIMDEUM multi-person households obtained with \ac{sdtw} clustering. \ac{sdtw}: (a) Family household cluster and (b) one- and two-person home cluster, all shown with barycenters $\mathbf{x^\ast}$, cluster means $\bm{\mu}$ and colors according to the resident number. Additionally, doughnut charts are shown in (b) depicting the \ac{sr} and \ac{er} in segmenting the daily patterns in one- and two-person homes vs. family homes. Histograms for the resident distribution within the clusters are depicted in (c), including translucent black bars for the total number of homes. Additionally, results with negative Silhouette coefficients ($S(\mathbf{y}_l)<0$) are highlighted. (d) Cluster analysis plots and (e) silhouette plots for $k=2$ for \ac{sdtw} clustering. (f) Comparison of clustering performance with respect to  \acp{sr}.}
	\label{fig:simdeum2_results}
\end{figure}

The cluster analysis is shown in Figure~\ref{fig:simdeum2_results}~(d) together with the individual silhouette coefficients for \ac{sdtw} clustering for $k=2$ (e).  The average silhouette value  is $\overline{S}=0.61$. 
All clustering algorithms identified  two distinct clusters. 
Clustering results for $k=2$ are depicted in Figure~\ref{fig:simdeum2_results}. 
Cluster $C_1$ contains mostly family households, $C_2$  one- and two-person homes.
It is assumed that the algorithm segments the daily patterns into family and non-family homes (one-person and two-person households). 
This consideration is taken into account to compute the success rate, which equals \ac{sr}$=97 \%$ (Figure~\ref{fig:simdeum2_results} (b)). 
A comparison with the benchmark algorithms shows again that \ac{sdtw} clustering has the highest \acp{sr} (see Figure~\ref{fig:simdeum2_results} (f)).
Figure~\ref{fig:simdeum2_results} (c) shows a histogram of the cluster members in dependency on the  resident numbers. 
The clusters are well separated for one and two persons as well as for four and more persons. 
Three-person households are present in both clusters with a much higher probability of being a member of $C_1$ (family households). 
\subsection{Measured dataset (Milford)}
\begin{figure}[htbp]
	\centering
	\includegraphics[width=1.0\linewidth]{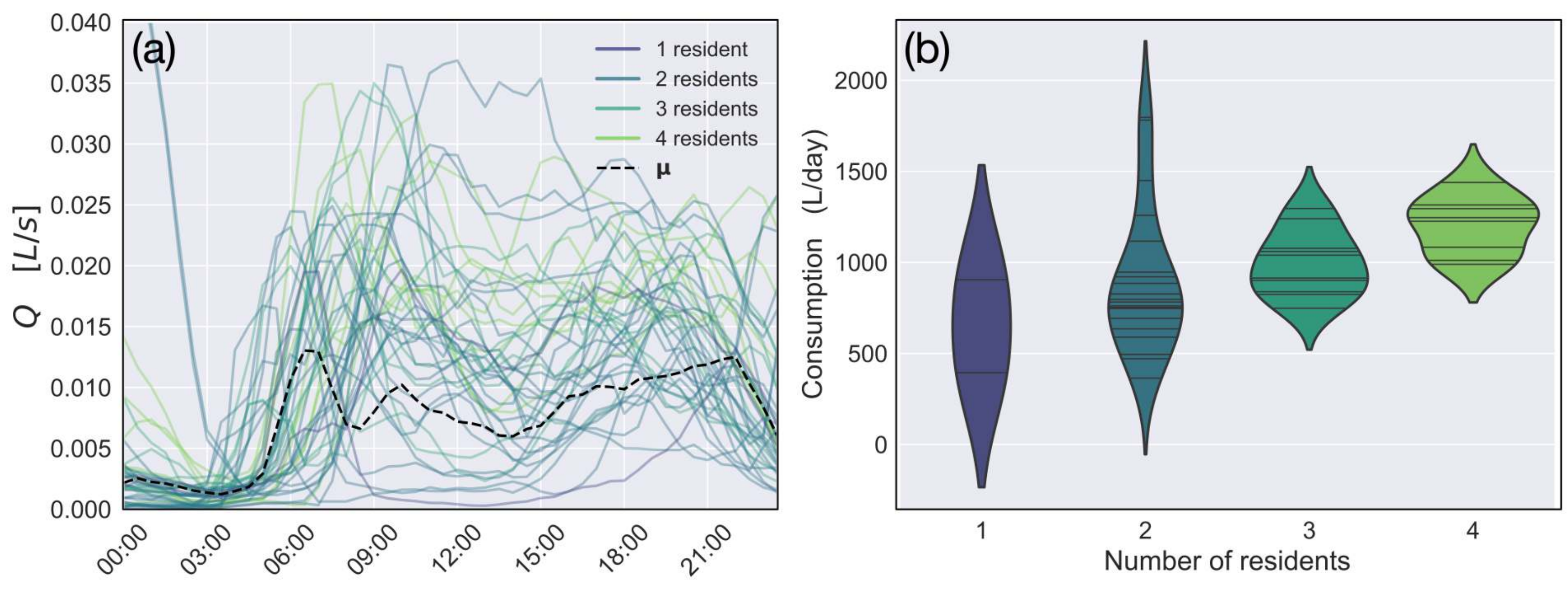}
	\caption{Representation of the Milford dataset consisting of 21 households and divided into weekday and weekends. (a): Daily weekday and weekend demand pattern of the households colored according to the number of residents. The mean $\bm{\mu}$ is depicted as a dashed black line. (b): Violin plots for the average daily consumption as a function of the resident number.}
	\label{fig:milford_dataset}
\end{figure}

The same technique are now applied to the Milford dataset.
Subsequently, a closer look at the clustering results and individual Silhouette coefficients is used to identify possible outliers within the clusters.
Figure~\ref{fig:milford_dataset} shows the dataset.
The increase in consumption by household size is not as prominent as for the SIMDEUM simulations (see Figure~\ref{fig:simdeum2_dataset}).
Furthermore, the variance is high leading to overlaps between households of all different resident numbers. 
Hence, the number of residents will not play a big role in segmenting the patterns as other information, e.g., the work schedules. 

\begin{figure}[htbp]
	\centering
	\includegraphics[width=1.0\linewidth]{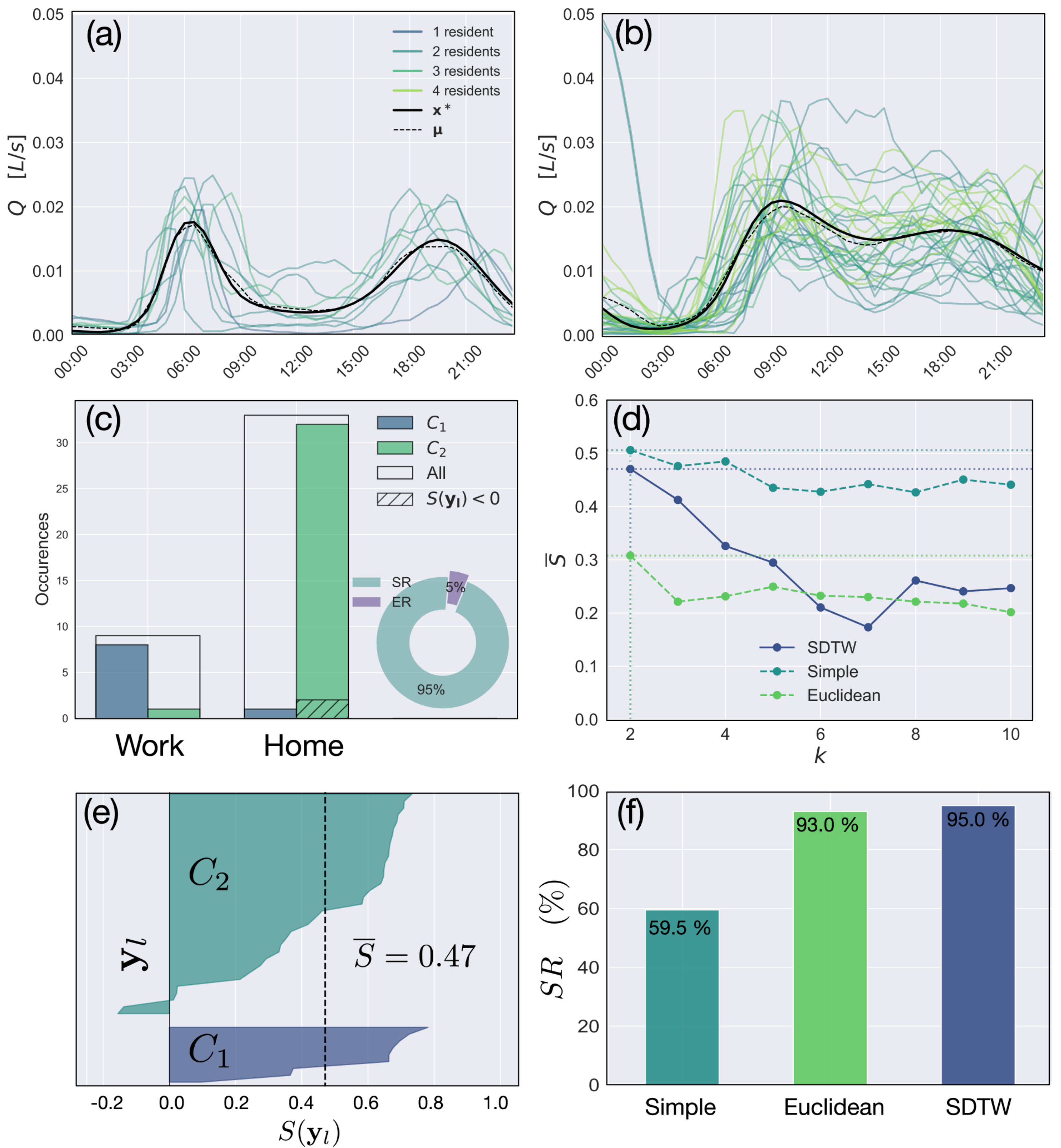}
	\caption{Clustering results for Milford dataset showing (a) the work cluster and (b) the home cluster, all shown with barycenters $\mathbf{x^\ast}$, cluster means $\bm{\mu}$ and colors according to the resident number. (c) Histograms for the resident distribution within the clusters are depicted in (c), including translucent black bars for the total number of homes. Results with negative Silhouette coefficients ($S(\mathbf{y}_l)<0$) are highlighted.  Additionally, doughnut charts are shown depicting the \ac{sr} and \ac{er} in segmenting the daily patterns in work and home patterns.  (d) Cluster analysis plots and (e) silhouette plots for $k=2$ for \ac{sdtw} clustering. (f) Comparison of clustering performance with respect to  \acp{sr}.}
	\label{fig:milford_results}
\end{figure}

\begin{figure}[htbp]
	\centering
	\includegraphics[width=0.7\linewidth]{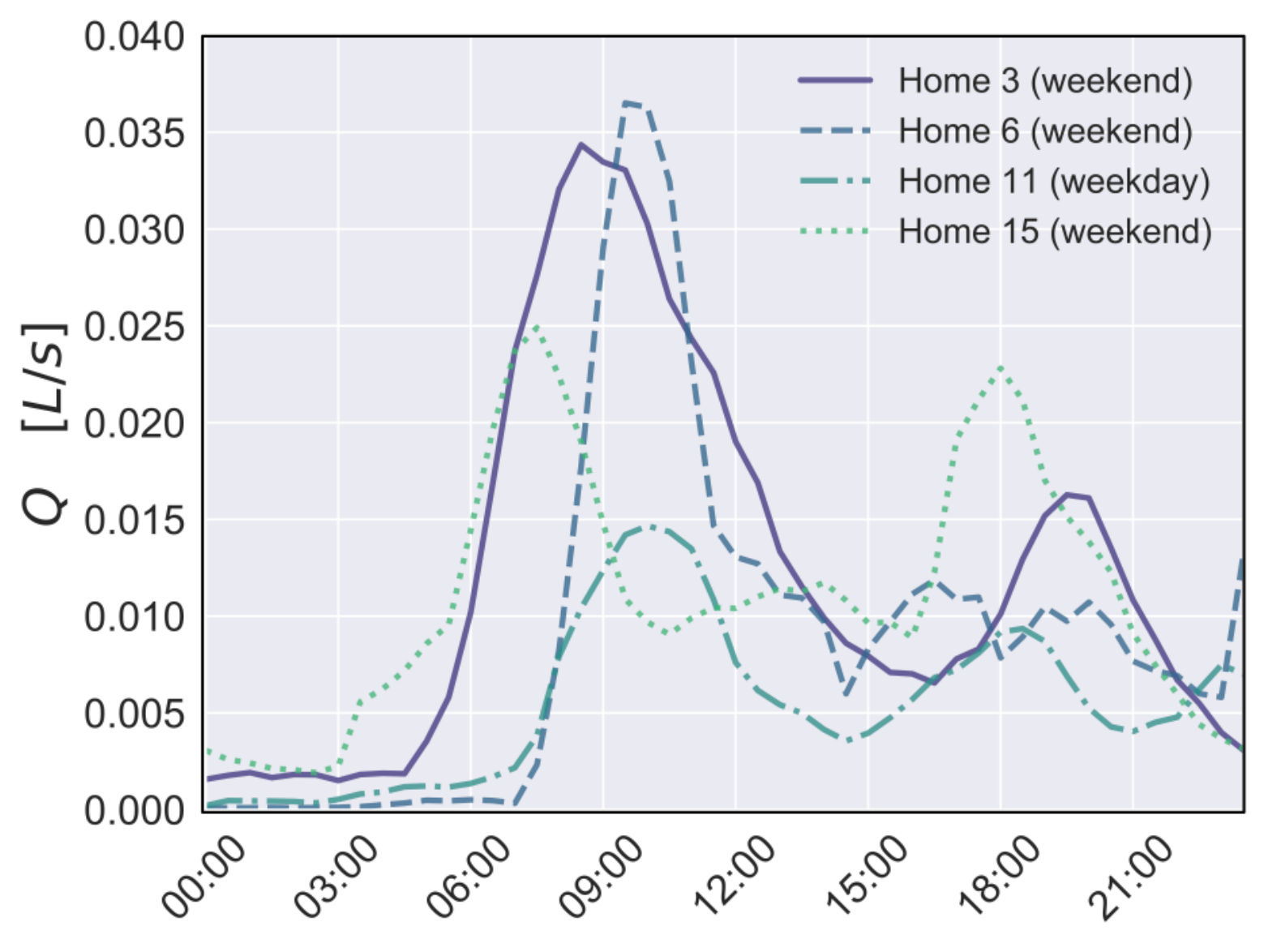}
	\caption{Special demand patterns identified in the Milford dataset.}
	\label{fig:milford_special_pattern}
\end{figure}

The results of the cluster analysis can be found in Figure \ref{fig:milford_results}~(d).
The average silhouette coefficients $\overline{S}$ show a very distinct maximum for $k$=2 for all clustering approaches; resulting in the assumption that  two distinct patterns are present in the dataset.
Figure~\ref{fig:milford_results} (a) to (c) present clustering results for  \ac{sdtw} for $k$=2. 
The clusters are not connected to the number of residents as for the SIMDEUM multi-person homes, but show to be dependent on the residents work schedules. 
The barycenters $\mathbf{x^\ast}$ clearly show that cluster $C_1$ (Figure~\ref{fig:milford_results} (a)) is the work cluster, $C_2$ represents the home cluster (Figure~\ref{fig:milford_results} (b)).  
Thus, the work schedules are identified as the most influential high-level information with a \ac{sr}=95 \%. 
\ac{sdtw} clustering results again in the highest \acp{sr} (see Figure~\ref{fig:milford_results} (f)). 
Additionally, Figure~\ref{fig:milford_results} (c) shows a histogram of the cluster members in dependency of their work schedules. 
The clusters are well separated in the work  and home cluster. 
Only two patterns are misclassified. 
These special cases are depicted in Figure~\ref{fig:milford_special_pattern}.
The weekday pattern of Home 11 has a different shape than all other patterns with three peaks. 
It is marked as a work pattern through expert opinion, but is a cluster member of the home cluster. 
The weekend pattern of Home 15 is the other way around: classified as work pattern, but marked as home pattern (Note that all weekend patterns are supposed to be home patterns). 
A closer look reveals a shape that is between the shape of work and home patterns, showing distinct morning and evening peaks but no prominent valley during the day. 
Furthermore, by looking at the negative Silhouette coefficients, two dissimilar patterns are identified (weekend patterns of Home 3 and 6). 
These patterns have a pre-eminent morning peak, while missing a peak in the evening. 

Additionally to the results presented in this section, further numerical studies on the same datasets can be found in the supplemental materials accompanying  this paper. The additional results contain comparisons between \ac{sdtw} algorithm with a clustering based on the original \ac{dtw} score, and time invariance robustness tests of Euclidean clustering compared to the \ac{sdtw} algorithm. Both tests were performed on the first dataset. In both cases, the \ac{sdtw} algorithm showed to be superior compared to \ac{dtw}  and Euclidean clustering. Moreover, the supplemental material contains more details on the benchmark tests comparing the performance of \ac{sdtw} clustering with the Simple as well as the Euclidean clustering algorithm.

\section{Discussion}
%
As outlined in the Introduction, the first question this study sought to determine was the number of distinct demand patterns in a specific \ac{swm} dataset.
Cluster analysis was used to answer this question. 
The \ac{sdtw} clustering method was able to identify the correct number of clusters ($k=2$) for the simulated single-person households  (Figure  \ref{fig:simdeum_cluster_analysis}), which correspond to the work schedules of the residents. 
Cluster analysis on the SIMDEUM multiple-person clearly revealed the presence of two different demand pattern types (Figure~\ref{fig:simdeum2_results}~(d) and \ref{fig:milford_results}~(d)).
The clusters of the SIMDEUM multi-person household are connected to the type of houses (family vs. one- and two person houses) or to the number of residents, respectively. Although the number of residents ranges from one to five, we assume that the household’s average consumption and its variance leads to substantial overlaps of patterns with different resident numbers, making it difficult for clustering algorithms to disaggregate the dataset in five separated groups.  It has to be noted that none of the benchmarking algorithm was able to identify more than two clusters in this dataset, as well. 
For the Milford dataset, two different clusters have been found, which are linked to the residents' work schedules.
By looking at the individual Silhouette coefficients in the Milford dataset, additional outlier patterns were identified with different shapes (e.g., a missing evening peak) (Figure \ref{fig:milford_special_pattern}; Home 3 and 6).
Note that the identification of outlier patterns can be done in an automatic way.
Benchmark tests with other clustering algorithms clearly showed a better performance of the \ac{sdtw} algorithm with respect to success rates over all datasets (see supplemental material for further details).
However, the better clustering performance of the \ac{sdtw} method comes with  higher computational times introduced by \ac{dtw} distance computations.
Nonetheless, the computation time grows only linearly with the length of the time series. 
For the data analyzed in the manuscript  the clustering analysis took less than a minute on a common personal computer. Therefore, computational complexity does not state a problem for real world applications.

The second question addressed the underlying high-level information responsible for the  different pattern shapes. 
For the SIMDEUM single-person household dataset, the differences in the patterns were caused by the residents' employment status (work, home). 
In this case, the shapes of the \ac{sdtw} barycenters can be intuitively linked with the employment characteristics, 
resembling qualitatively better the expected behavior (e.g., work patterns have low valleys in consumption while the residents are at work). 
For the SIMDEUM multi-person households, the clustering approach resulted in clusters based on household type (one- and two-person versus family) with a high accuracy of 97 \%. 
For the Milford dataset, barycenters  were  linked  to the residents' work schedules. 
Application of the clustering algorithms resulted in barycenters of a home cluster with high consumption during the day and a work cluster in which the consumption is low during work hours.
Reasons for the low but non-zero consumption of the work cluster could be (i) that the households are multiple-person households in contrast to single-person households in the first SIMDEUM dataset and, hence, some of the inhabitants stay at home during the day, or (ii) that the inhabitants have different daily schedules on different days of the week, e.g., four-day jobs.
In summary, it can be said that the out-of-home activities (work, school, \ldots) and the number of household residents are the most important  high-level information revealed by the automatic clustering algorithm.

In future, we plan to focus on analyzing more complex datasets (i) with other important high-level information of (e.g., household income, age, and gender distribution), (ii) from different countries, (iii) with different end-use devices or (iv) disaggregated by end-uses.
Besides the clustering of consumption patterns, the proposed method is additionally valuable for (i)  finding outlier patterns between different customers (e.g. unusual high water consumption) or (ii) identifying changes in patterns over time (e.g. changing daily routines caused by illness or unemployment).
The next research steps will concentrate on parameterizing stochastic end-use models based on this approach to provide more realistic demand simulation tools.
\section{Conclusion}
Since water demand is shaped by socio-economic characteristics, knowledge on these characteristics and how they are connected with the dynamics of water consumption is highly valuable for \ac{wds} modeling. 
This work shows how data science algorithms  can be used to link \ac{swm} data to high-level information.
The novel \ac{sdtw} clustering technique is capable of finding similarities in daily demand patterns even when they have similar features shifted in time. 
In this manuscript, the technique is tested on simulated and measured \ac{swm} datasets.
It is shown for the dataset where the ground truth is known that \ac{sdtw}  clustering is able to classify patterns accurately as well as to identify the correct number of patterns.
It is shown that \ac{sdtw} clustering outperforms commonly used clustering algorithms (e.g. Euclidean clustering).
Furthermore, the shape of the cluster's barycenters can be linked to user characteristics. 
The employment status and the number of household residents is identified as the most important underlying high-level information.
Additionally, the methodology presented in this work can be used to identify outlier within demand patterns.

Generally, the findings of this study clearly demonstrated that socio-economic characteristics manifest themselves in the shapes of water usage patterns and, hence, these characteristics can be identified from the datasets through the proposed clustering approaches.
Since demand patterns can be linked to high-level information, this information can be used to infer and simulate water consumption at unmeasured points  in a \ac{wds}, either by using directly the daily demand patterns in hydraulic simulation software, or by using a SIMDEUM model parameterized by customers' socio-economic data. For example, in the Netherlands, socio-economic data are freely available at a neighborhood (post-code) level from the national statistical agency. 
This offers the opportunity of complementing data-gaps in hydraulic models and, hence, the possibility of reducing model uncertainties.
%
%
%
\appendix
\section{Data Availability Statement}
Some or all data, models, or code generated or used during the study are available in a repository or online in accordance with funder data retention policies. (\url{https://github.com/steffelbauer/swm_sdtw})
\section{Acknowledgments}
This project has received funding from the European Union's Horizon 2020 research and innovation programme under the Marie Sk\l odowska-Curie grant agreement No 707404. The opinions expressed in this document reflect only the author's view. The European Commission is not responsible for any use that may be made of the information it contains. The authors want to thank Professor Eamonn Keogh for pointing out that dynamic time warping is only a distance measure and not a metric.
\bibliographystyle{abbrvnat}
\bibliography{references}  

\begin{thebibliography}{45}
\providecommand{\natexlab}[1]{#1}
\providecommand{\url}[1]{\texttt{#1}}
\expandafter\ifx\csname urlstyle\endcsname\relax
  \providecommand{\doi}[1]{doi: #1}\else
  \providecommand{\doi}{doi: \begingroup \urlstyle{rm}\Url}\fi

\bibitem[Aksela and Aksela(2011)]{Aksela2011}
K.~Aksela and M.~Aksela.
\newblock {Demand Estimation with Automated Meter Reading in a Distribution
  Network}.
\newblock \emph{Journal of Water Resources Planning and Management},
  137\penalty0 (5):\penalty0 456--467, 2011.
\newblock \doi{10.1061/(ASCE)WR.1943-5452.0000131}.

\bibitem[Alvisi et~al.(2007)Alvisi, Franchini, and Marinelli]{Alvisi2007}
S.~Alvisi, M.~Franchini, and A.~Marinelli.
\newblock {A short-term, pattern-based model for water-demand forecasting}.
\newblock \emph{Journal of Hydroinformatics}, 9\penalty0 (1):\penalty0 39--50,
  2007.
\newblock ISSN 14647141.
\newblock \doi{10.2166/hydro.2006.016}.

\bibitem[Arthur and Vassilvitskii(2007)]{Arthur2007}
D.~Arthur and S.~Vassilvitskii.
\newblock K-means++: The advantages of careful seeding.
\newblock In \emph{{Proceedings of the Eighteenth Annual ACM-SIAM Symposium on
  Discrete Algorithms}}, SODA '07, pages 1027--1035, Philadelphia, PA, USA,
  2007. Society for Industrial and Applied Mathematics.
\newblock \doi{10.1145/1283383.1283494}.

\bibitem[Blokker et~al.(2008)Blokker, Vreeburg, Buchberger, and van
  Dijk]{Blokker2008}
E.~J.~M. Blokker, J.~H.~G. Vreeburg, S.~G. Buchberger, and J.~C. van Dijk.
\newblock {Importance of demand modelling in network water quality models: a
  review}.
\newblock \emph{Drinking Water Engineering and Science}, 1\penalty0
  (1):\penalty0 27--38, 2008.
\newblock \doi{10.5194/dwes-1-27-2008}.

\bibitem[Blokker et~al.(2010)Blokker, Vreeburg, and van Dijk]{Blokker2010}
E.~J.~M. Blokker, J.~H.~G. Vreeburg, and J.~C. van Dijk.
\newblock {Simulating Residential Water Demand with a Stochastic End-Use
  Model}.
\newblock \emph{Journal of Water Resources Planning and Management},
  136\penalty0 (1):\penalty0 19--26, 2010.
\newblock \doi{10.1061/(ASCE)WR.1943-5452.0000146.}

\bibitem[Boyle et~al.(2013)Boyle, Giurco, Mukheibir, Liu, Moy, White, and
  Stewart]{Boyle2013}
T.~Boyle, D.~Giurco, P.~Mukheibir, A.~Liu, C.~Moy, S.~White, and R.~Stewart.
\newblock {Intelligent metering for urban water: A review}.
\newblock \emph{Water (Switzerland)}, 5\penalty0 (3):\penalty0 1052--1081,
  2013.
\newblock \doi{10.3390/w5031052}.

\bibitem[Buchberger et~al.(2003)Buchberger, Carter, Lee, and
  Schade]{Buchberger2003}
S.~G. Buchberger, J.~Carter, Y.~Lee, and T.~G. Schade.
\newblock \emph{{Random demands, travel times, and water quality in deadends}}.
\newblock AWWA Research Foundation, Denver, Colorado, USA, 2003.

\bibitem[Candelieri(2017)]{Candelieri2017}
A.~Candelieri.
\newblock {Clustering and support vector regression for water demand
  forecasting and anomaly detection}.
\newblock \emph{Water (Switzerland)}, 9\penalty0 (3), 2017.
\newblock \doi{10.3390/w9030224}.

\bibitem[Cardell-Oliver(2013{\natexlab{a}})]{Cardell-Oliver2013}
R.~Cardell-Oliver.
\newblock {Water use signature patterns for analyzing household consumption
  using medium resolution meter data}.
\newblock \emph{Water Resources Research}, 49\penalty0 (12):\penalty0
  8589--8599, 2013{\natexlab{a}}.
\newblock ISSN 00431397.
\newblock \doi{10.1002/2013WR014458}.

\bibitem[Cardell-Oliver(2013{\natexlab{b}})]{Cardell-Oliver2013a}
R.~Cardell-Oliver.
\newblock {Discovering water use activities for smart metering}.
\newblock In \emph{2013 IEEE Eighth International Conference on Intelligent
  Sensors, Sensor Networks and Information Processing}, pages 171--176. IEEE,
  2013{\natexlab{b}}.
\newblock ISBN 978-1-4673-5501-8.
\newblock \doi{10.1109/ISSNIP.2013.6529784}.

\bibitem[Cardell-Oliver et~al.(2016)Cardell-Oliver, Wang, and
  Gigney]{Cardell-Oliver2016}
R.~Cardell-Oliver, J.~Wang, and H.~Gigney.
\newblock {Smart Meter Analytics to Pinpoint Opportunities for Reducing
  Household Water Use}.
\newblock \emph{Journal of Water Resources Planning and Management},
  142\penalty0 (6):\penalty0 04016007, 2016.
\newblock ISSN 0733-9496.
\newblock \doi{10.1061/(ASCE)WR.1943-5452.0000634}.

\bibitem[Cheifetz et~al.(2017)Cheifetz, Noumir, Sam{\'{e}}, Sandraz,
  F{\'{e}}liers, and Heim]{Cheifetz2017}
N.~Cheifetz, Z.~Noumir, A.~Sam{\'{e}}, A.-C. Sandraz, C.~F{\'{e}}liers, and
  V.~Heim.
\newblock {Modeling and clustering water demand patterns from real-world smart
  meter data}.
\newblock \emph{Drinking Water Engineering and Science}, 10\penalty0
  (2):\penalty0 75--82, 2017.
\newblock \doi{10.5194/dwes-10-75-2017}.

\bibitem[Cominola et~al.(2015)Cominola, Giuliani, Piga, Castelletti, and
  Rizzoli]{Cominola2015}
A.~Cominola, M.~Giuliani, D.~Piga, A.~Castelletti, and A.~Rizzoli.
\newblock {Benefits and challenges of using smart meters for advancing
  residential water demand modeling and management: A review}.
\newblock \emph{Environmental Modelling {\&} Software}, 72:\penalty0 198--214,
  2015.
\newblock \doi{10.1016/J.ENVSOFT.2015.07.012}.

\bibitem[Cominola et~al.(2018)Cominola, Giuliani, Castelletti, Rosenberg, and
  Abdallah]{Cominola2018}
A.~Cominola, M.~Giuliani, A.~Castelletti, D.~Rosenberg, and A.~Abdallah.
\newblock {Implications of data sampling resolution on water use simulation,
  end-use disaggregation, and demand management}.
\newblock \emph{Environmental Modelling {\&} Software}, 102:\penalty0 199--212,
  2018.
\newblock ISSN 1364-8152.
\newblock \doi{10.1016/J.ENVSOFT.2017.11.022}.

\bibitem[Cominola et~al.(2019)Cominola, Nguyen, Giuliani, Stewart, Maier, and
  Castelletti]{Cominola2019}
A.~Cominola, K.~Nguyen, M.~Giuliani, R.~A. Stewart, H.~R. Maier, and
  A.~Castelletti.
\newblock {Data mining to uncover heterogeneous water use behaviors from smart
  meter data}.
\newblock \emph{Water Resources Research}, page 2019WR024897, 2019.
\newblock ISSN 0043-1397.
\newblock \doi{10.1029/2019WR024897}.

\bibitem[Cuturi and Blondel(2017)]{Cuturi2017}
M.~Cuturi and M.~Blondel.
\newblock Soft-{DTW}: a differentiable loss function for time-series.
\newblock In D.~Precup and Y.~W. Teh, editors, \emph{Proceedings of the 34th
  International Conference on Machine Learning}, volume~70 of
  \emph{{Proceedings of Machine Learning Research}}, pages 894--903,
  International Convention Centre, Sydney, Australia, 2017. PMLR.

\bibitem[D{\'{i}}az and Gonz{\'{a}}lez(2020)]{Diaz2020}
S.~D{\'{i}}az and J.~Gonz{\'{a}}lez.
\newblock {Analytical Stochastic Microcomponent Modeling Approach to Assess
  Network Spatial Scale Effects in Water Supply Systems}.
\newblock \emph{Journal of Water Resources Planning and Management},
  146\penalty0 (8):\penalty0 04020065, 2020.
\newblock ISSN 0733-9496.
\newblock \doi{10.1061/(ASCE)WR.1943-5452.0001237}.

\bibitem[D{\"{u}}rrenmatt(2011)]{Durrenmatt2011}
D.~D{\"{u}}rrenmatt.
\newblock \emph{{Data Mining and Data-Driven Modeling Approaches To Support
  Wastewater Treatment Plant Operation}}.
\newblock Doctoral thesis, ETH Z{\"{u}}rich, Switzerland, 2011.

\bibitem[D{\"u}rrenmatt et~al.(2013)D{\"u}rrenmatt, Giudice, and
  Rieckermann]{Durrenmatt2013}
D.~J. D{\"u}rrenmatt, D.~D. Giudice, and J.~Rieckermann.
\newblock {Dynamic time warping improves sewer flow monitoring}.
\newblock \emph{Water Research}, 47\penalty0 (11):\penalty0 3803--3816, 2013.
\newblock \doi{10.1016/J.WATRES.2013.03.051}.

\bibitem[Espinoza et~al.(2005)Espinoza, Joye, Belmans, and {De
  Moor}]{Espinoza2005}
M.~Espinoza, C.~Joye, R.~Belmans, and B.~{De Moor}.
\newblock {Short-term load forecasting, profile identification, and customer
  segmentation: A methodology based on periodic time series}.
\newblock \emph{IEEE Transactions on Power Systems}, 20\penalty0 (3):\penalty0
  1622--1630, 2005.
\newblock ISSN 08858950.
\newblock \doi{10.1109/TPWRS.2005.852123}.

\bibitem[Garcia et~al.(2015)Garcia, {Gonz{\'{a}}lez Vidal}, Quevedo, Puig, and
  Saludes]{Garcia2015}
D.~Garcia, D.~{Gonz{\'{a}}lez Vidal}, J.~Quevedo, V.~Puig, and J.~Saludes.
\newblock {Water Demand Estimation and Outlier Detection from Smart Meter Data
  Using Classification and Big Data Methods}.
\newblock \emph{{2nd New Developments in IT {\&} Water Conference, Rotterdam,
  Netherlands}}, pages 1--8, 2015.

\bibitem[Gurung et~al.(2014)Gurung, Stewart, Sharma, and Beal]{Gurung2014}
T.~R. Gurung, R.~A. Stewart, A.~K. Sharma, and C.~D. Beal.
\newblock {Smart meters for enhanced water supply network modelling and
  infrastructure planning}.
\newblock \emph{Resources, Conservation and Recycling}, 90:\penalty0 34--50,
  2014.
\newblock \doi{10.1016/j.resconrec.2014.06.005}.

\bibitem[{Haestad Methods}(2003)]{Haestad2003}
{Haestad Methods}.
\newblock \emph{{Advanced water distribution modeling and management}}.
\newblock Haestead Press, Waterbury, CT, 1 edition, 2003.
\newblock ISBN 0-9714141-2-2.

\bibitem[Huang et~al.(2018)Huang, Zhu, Hou, Chen, Xiao, Yu, Zhang, and
  Zhang]{Huang2018}
P.~Huang, N.~Zhu, D.~Hou, J.~Chen, Y.~Xiao, J.~Yu, G.~Zhang, and H.~Zhang.
\newblock {Real-time burst detection in District Metering Areas in water
  distribution system based on patterns of water demand with supervised
  learning}.
\newblock \emph{Water (Switzerland)}, 10\penalty0 (12):\penalty0 1765, 2018.
\newblock \doi{10.3390/w10121765}.

\bibitem[Hutton et~al.(2014)Hutton, Kapelan, Vamvakeridou-Lyroudia, and
  Savi{\'{c}}]{Hutton2012}
C.~J. Hutton, Z.~Kapelan, L.~Vamvakeridou-Lyroudia, and D.~A. Savi{\'{c}}.
\newblock {Dealing with Uncertainty in Water Distribution System Models: A
  Framework for Real-Time Modeling and Data Assimilation}.
\newblock \emph{Journal of Water Resources Planning and Management},
  140\penalty0 (2):\penalty0 169--183, 2014.
\newblock \doi{10.1061/(asce)wr.1943-5452.0000325}.

\bibitem[Kwac et~al.(2014)Kwac, Flora, and Rajagopal]{Kwac2014}
J.~Kwac, J.~Flora, and R.~Rajagopal.
\newblock {Household energy consumption segmentation using hourly data}.
\newblock \emph{IEEE Transactions on Smart Grid}, 5\penalty0 (1):\penalty0
  420--430, 2014.
\newblock ISSN 19493053.
\newblock \doi{10.1109/TSG.2013.2278477}.

\bibitem[Lin et~al.(2012)Lin, Williamson, Borne, and DeBarr]{Lin2012}
J.~Lin, S.~Williamson, K.~Borne, and D.~DeBarr.
\newblock {Pattern Recognition in Time Series}.
\newblock \emph{Advances in Machine Learning and Data Mining for Astronomy},
  1:\penalty0 617--645, 2012.
\newblock \doi{10.1201/b11822-36}.

\bibitem[Lloyd(2006)]{Lloyd1982}
S.~Lloyd.
\newblock {Least Squares Quantization in PCM}.
\newblock \emph{IEEE Trans. Inf. Theor.}, 28\penalty0 (2):\penalty0 129--137,
  2006.
\newblock ISSN 0018-9448.
\newblock \doi{10.1109/TIT.1982.1056489}.

\bibitem[McKenna et~al.(2014)McKenna, Fusco, and Eck]{McKenna2014}
S.~A. McKenna, F.~Fusco, and B.~J. Eck.
\newblock {Water demand pattern classification from smart meter data}.
\newblock \emph{Procedia Engineering}, 70:\penalty0 1121--1130, 2014.
\newblock \doi{10.1016/j.proeng.2014.02.124}.

\bibitem[Monks et~al.(2019)Monks, Stewart, Sahin, and Keller]{Monks2019}
I.~Monks, R.~A. Stewart, O.~Sahin, and R.~Keller.
\newblock {Revealing Unreported Benefits of Digital Water Metering: Literature
  Review and Expert Opinions}.
\newblock \emph{Water}, 11\penalty0 (4):\penalty0 838, 2019.
\newblock ISSN 2073-4441.
\newblock \doi{10.3390/w11040838}.

\bibitem[Mounce et~al.(2016)Mounce, Furnass, Goya, Hawkins, and
  Boxall]{Mounce2016}
S.~R. Mounce, W.~R. Furnass, E.~Goya, M.~Hawkins, and J.~B. Boxall.
\newblock {Clustering and classification of aggregated smart meter data to
  better understand how demand patterns relate to customer type}.
\newblock In \emph{{Proceedings of the 14th International Conference of
  Computing and Control for the Water Industry -- CCWI 2016}}, pages 1--9,
  Amsterdam, Netherlands, 2016.

\bibitem[Nambi et~al.(2016)Nambi, Pournaras, and Prasad]{Nambi2016}
S.~N. Nambi, E.~Pournaras, and R.~V. Prasad.
\newblock {Temporal Self-Regulation of Energy Demand}.
\newblock \emph{IEEE Transactions on Industrial Informatics}, 12\penalty0
  (3):\penalty0 1196--1205, 2016.
\newblock ISSN 15513203.
\newblock \doi{10.1109/TII.2016.2554519}.

\bibitem[Nguyen et~al.(2014)Nguyen, Stewart, and Zhang]{Nguyen2014}
K.~A. Nguyen, R.~A. Stewart, and H.~Zhang.
\newblock {An autonomous and intelligent expert system for residential water
  end-use classification}.
\newblock \emph{Expert Systems with Applications}, 41\penalty0 (2):\penalty0
  342--356, 2014.
\newblock \doi{10.1016/j.eswa.2013.07.049}.

\bibitem[Nguyen et~al.(2018)Nguyen, Stewart, Zhang, Sahin, and
  Siriwardene]{Nguyen2018}
K.~A. Nguyen, R.~A. Stewart, H.~Zhang, O.~Sahin, and N.~Siriwardene.
\newblock {Re-engineering traditional urban water management practices with
  smart metering and informatics}.
\newblock \emph{Environmental Modelling and Software}, 101:\penalty0 256--267,
  2018.
\newblock \doi{10.1016/j.envsoft.2017.12.015}.

\bibitem[Nizar and Dong(2009)]{Nizar2009}
A.~H. Nizar and Z.~Y. Dong.
\newblock {Identification and detection of electricity customer behaviour
  irregularities}.
\newblock In \emph{2009 IEEE/PES Power Systems Conference and Exposition, PSCE
  2009}, 2009.
\newblock ISBN 9781424438112.
\newblock \doi{10.1109/PSCE.2009.4840253}.

\bibitem[Nocedal and Wright(2006)]{Nocedal2006}
J.~Nocedal and S.~J. Wright.
\newblock \emph{Numerical Optimization}.
\newblock Springer, New York, NY, USA, 2 edition, 2006.

\bibitem[Rousseeuw(1987)]{Rousseeuw1987}
P.~Rousseeuw.
\newblock Silhouettes: A graphical aid to the interpretation and validation of
  cluster analysis.
\newblock \emph{J. Comput. Appl. Math.}, 20\penalty0 (1):\penalty0 53--65,
  1987.
\newblock \doi{10.1016/0377-0427(87)90125-7}.

\bibitem[Sakoe and Chiba(1978)]{Sakoe1978}
H.~Sakoe and S.~Chiba.
\newblock {Dynamic programming algorithm optimization for spoken word
  recognition}.
\newblock \emph{IEEE Transactions on Acoustics, Speech, and Signal Processing},
  26\penalty0 (1):\penalty0 43--49, 1978.
\newblock \doi{10.1109/TASSP.1978.1163055}.

\bibitem[Shafiee et~al.(2020)Shafiee, Rasekh, Sela, Preis, Rasekh, Sela, Asce,
  and Preis]{Shafiee2020}
M.~E. Shafiee, A.~Rasekh, L.~Sela, A.~Preis, .~A. Rasekh, L.~Sela, M.~Asce, and
  A.~Preis.
\newblock {Streaming Smart Meter Data Integration to Enable Dynamic Demand
  Assignment for Real-Time Hydraulic Simulation}.
\newblock \emph{Journal of Water Resources Planning and Management},
  146\penalty0 (6):\penalty0 06020008, 2020.
\newblock ISSN 0733-9496.
\newblock \doi{10.1061/(ASCE)WR.1943-5452.0001221}.

\bibitem[Shumway and Stoffer(2010)]{Shumway2010}
R.~H. Shumway and D.~S. Stoffer.
\newblock \emph{{Time Series Analysis and Its Applications With R Examples}}.
\newblock Springer International Publishing, Basel, Switzerland, 4 edition,
  2010.
\newblock \doi{10.1007/978-3-319-52452-8}.

\bibitem[Stewart et~al.(2018)Stewart, Nguyen, Beal, Zhang, Sahin, Bertone,
  Vieira, Castelletti, Cominola, Giuliani, Giurco, Blumenstein, Turner, Liu,
  Kenway, Savi{\'{c}}, Makropoulos, and Kossieris]{Stewart2018}
R.~A. Stewart, K.~Nguyen, C.~Beal, H.~Zhang, O.~Sahin, E.~Bertone, A.~S.
  Vieira, A.~Castelletti, A.~Cominola, M.~Giuliani, D.~Giurco, M.~Blumenstein,
  A.~Turner, A.~Liu, S.~Kenway, D.~A. Savi{\'{c}}, C.~Makropoulos, and
  P.~Kossieris.
\newblock {Integrated intelligent water-energy metering systems and
  informatics: Visioning a digital multi-utility service provider}.
\newblock \emph{Environmental Modelling {\&} Software}, 105:\penalty0 94--117,
  2018.
\newblock \doi{10.1016/J.ENVSOFT.2018.03.006}.

\bibitem[Wang et~al.(2015)Wang, Cardell-Oliver, and Liu]{Wang2015}
J.~Wang, R.~Cardell-Oliver, and W.~Liu.
\newblock {Efficient discovery of recurrent routine behaviours in smart meter
  time series by growing subsequences}.
\newblock In \emph{Lecture Notes in Computer Science (including subseries
  Lecture Notes in Artificial Intelligence and Lecture Notes in
  Bioinformatics)}, volume 9078, pages 522--533. Springer Verlag, 2015.
\newblock ISBN 9783319180311.
\newblock \doi{10.1007/978-3-319-18032-8_41}.

\bibitem[Witten et~al.(2011)Witten, Frank, and Hall]{Witten2011}
I.~H. Witten, E.~Frank, and M.~A. Hall.
\newblock \emph{Data Mining: Practical Machine Learning Tools and Techniques}.
\newblock Morgan Kaufmann Publishers Inc., San Francisco, CA, USA, 3rd edition,
  2011.
\newblock ISBN 0123748569, 9780123748560.

\bibitem[Yang et~al.(2018)Yang, Zhang, Stewart, and Nguyen]{Yang2018}
A.~Yang, H.~Zhang, R.~A. Stewart, and K.~Nguyen.
\newblock {Enhancing residential water end use pattern recognition accuracy
  using self-organizing maps and K-means clustering techniques: Autoflow v3.1}.
\newblock \emph{Water (Switzerland)}, 10\penalty0 (9), 2018.
\newblock \doi{10.3390/w10091221}.

\bibitem[Zhou et~al.(2018)Zhou, Xu, Xin, Yan, and Tao]{Zhou2018}
X.~Zhou, W.~Xu, K.~Xin, H.~Yan, and T.~Tao.
\newblock {Self-Adaptive Calibration of Real-Time Demand and Roughness of Water
  Distribution Systems}.
\newblock \emph{Water Resources Research}, 54\penalty0 (8):\penalty0
  5536--5550, 2018.
\newblock \doi{10.1029/2017WR022147}.

\end{thebibliography}





%
\begin{acronym}
	\acro{wef}[WEF]{Water, Energy and Food}
	\acro{dtw}[DTW]{Dynamic Time Warping}
	\acro{er}[ER]{Error Rate}
	\acro{fn}[FN]{False Negative}
	\acro{fp}[FP]{False Positive}
	\acro{prp}[PRP]{Poisson Rectangular Pulse}
	\acro{sdtw}[SDTW]{Soft Dynamic Time Warping}
	\acro{sr}[SR]{Success Rate}
	\acro{swm}[SWM]{Smart Water Meter}
	\acro{tn}[TN]{True Negative}
	\acro{tp}[TP]{True Positive}
	\acro{wds}[WDS]{Water Distribution System}
	\acro{wu}[WU]{Water Utility}
\end{acronym}

\end{document}